\documentclass[letterpaper]{article} 
\usepackage{aaai25}  
\usepackage{times}  
\usepackage{helvet}  
\usepackage{courier}  
\usepackage[hyphens]{url}  
\usepackage{graphicx} 
\usepackage{natbib}  
\usepackage{caption} 
\usepackage{algorithm}
\usepackage{algorithmic}
\usepackage{amsmath}
\usepackage{booktabs}
\usepackage{tabularx}
\usepackage{svg}
\usepackage{url}

\usepackage{newfloat}
\usepackage{listings}
\DeclareCaptionStyle{ruled}{labelfont=normalfont,labelsep=colon,strut=off} 
\floatstyle{ruled}
\newfloat{listing}{tb}{lst}{}
\floatname{listing}{Listing}
\nocopyright

\title{Beyond KV Caching: Shared Attention for Efficient LLMs}
\author {
    Liao Bingli\textsuperscript{\rm 1},
    Danilo Vasconcellos Vargas\textsuperscript{\rm 1}
}
\affiliations {
    \textsuperscript{\rm 1}Kyushu University, Fukuoka, Japan\\
    \{liao.bingli.734@s, vargas@inf\}.kyushu-u.ac.jp
}

\begin{document}

\maketitle

\begin{abstract}
The efficiency of large language models (LLMs) remains a critical challenge, particularly in contexts where computational resources are limited. Traditional attention mechanisms in these models, while powerful, require significant computational and memory resources due to the necessity of recalculating and storing attention weights across different layers. This paper introduces a novel Shared Attention (SA) mechanism, designed to enhance the efficiency of LLMs by directly sharing computed attention weights across multiple layers. Unlike previous methods that focus on sharing intermediate Key-Value (KV) caches, our approach utilizes the isotropic tendencies of attention distributions observed in advanced LLMs post-pretraining to reduce both the computational flops and the size of the KV cache required during inference. We empirically demonstrate that implementing SA across various LLMs results in minimal accuracy loss on standard benchmarks. Our findings suggest that SA not only conserves computational resources but also maintains robust model performance, thereby facilitating the deployment of more efficient LLMs in resource-constrained environments. Code: \url{https://github.com/metacarbon/shareAtt}
\end{abstract}

%

\section{Introduction}

The rapid growth of large language models (LLM) has brought forth significant challenges in terms of computational and memory efficiency during inference. Traditional approaches, such as Multi-Query Attention (MQA) \cite{shazeer2019fast} and Grouped-Query Attention (GQA) \cite{ainslie2023gqa}, have made strides in reducing the key-value (KV) cache size by sharing keys and values across multiple heads within a layer. More recently, Cross-Layer Attention (CLA) has extended this concept by sharing keys and values across adjacent layers, further reducing memory requirements without substantially impacting model performance \cite{brandon2024reducing}. Despite these advancements, the need for more efficient methods continues to grow, particularly as models scale and are deployed in resource-constrained environments.

In this paper, we introduce a novel method termed Shared Attention (SA), which significantly reduces the KV cache requirements and computational load during inference for LLMs. Unlike previous methods that focused on sharing KV caches either within the same layer or between adjacent layers, our approach inspired by the inherent similarity of attention weights distribution across layers, and sharing these weights directly could further reduce the need for repeated key and value computations. This innovative approach not only reduces the KV cache size but also circumvents the need for the computationally expensive softmax operation, leading to a more efficient inference process.

The key contributions of our work are summarized as follows:

\begin{enumerate}
    \item We propose a novel Shared Attention mechanism that reduces computational and memory overhead by directly sharing pre-computed attention weights across multiple layers in LLMs.
    \item We empirically validate the effectiveness of Shared Attention by implementing it across various benchmarks and demonstrate that it achieves comparable accuracy.
    \item Our analysis of attention isotropy across pretrained LLMs provides insights into how attention mechanisms stabilize and become more uniform across layers as training progresses. This understanding informs the optimal layer ranges for applying Shared Attention.
\end{enumerate}

\section{Shared Attention}
In this section we demonstrate motivation, Shared Attention (SA) method, and the comparison to existed KV-sharing mechanisms.

\begin{figure*}[ht]
    \centering
    \includegraphics[width=0.8\textwidth]{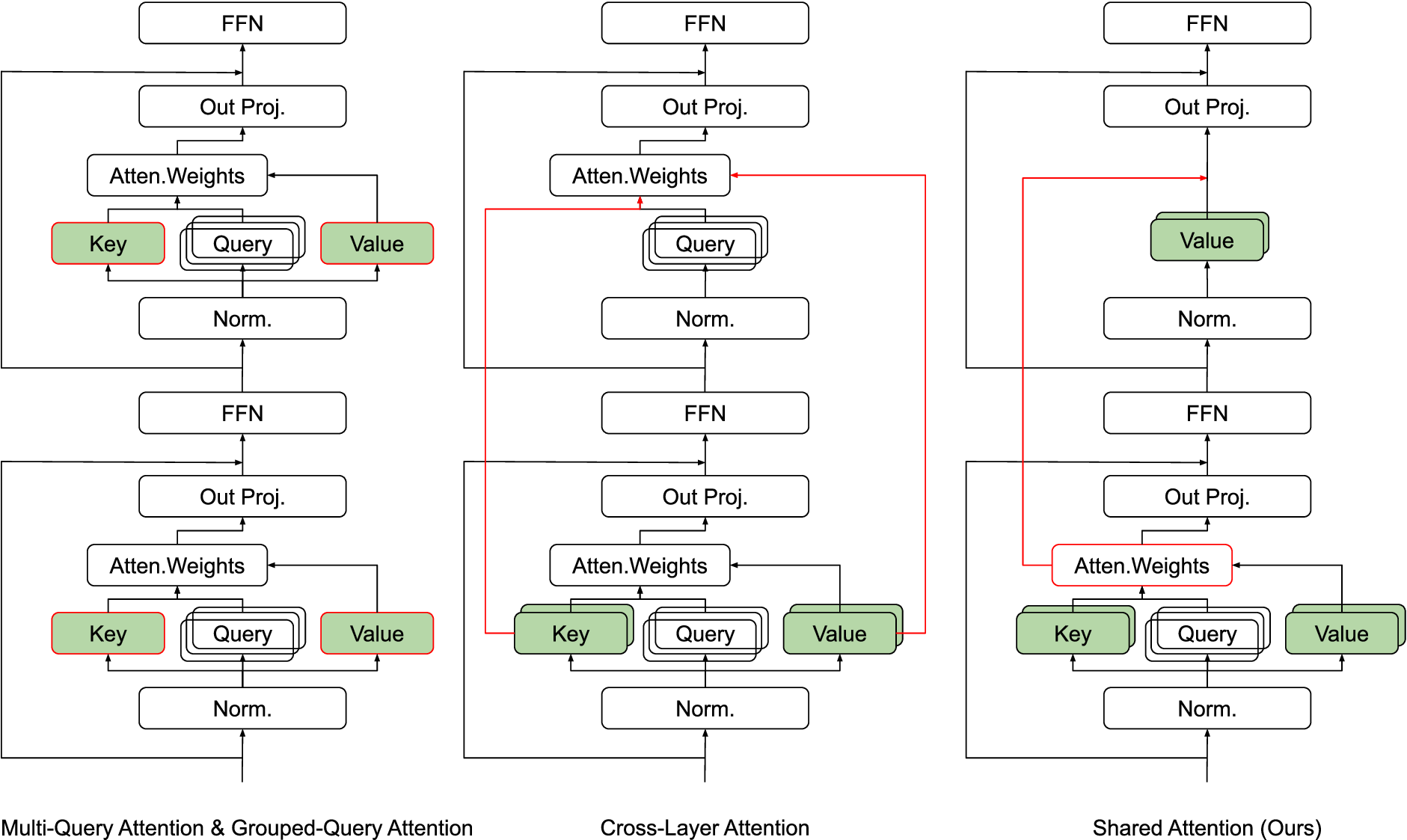}
    \caption{Illustration of various sharing algorithms. The MQA and GQA methods share the Key and Value caches with the Query within the same layer to reduce memory usage. The CLA method extends this by sharing the Key and Value caches across different layers. Our method, Shared Attention, advances this concept further by sharing the attention weights across multiple layers.}
    \label{fig:SA_CLA_Comparision}
\end{figure*}

\subsection{Motivation}
The self-attention mechanism in transformer models is typically defined as $\text{softmax}(\frac{QK^T}{\sqrt{d}})V$, where $Q$, $K$, and $V$ represent the query, key, and value matrices respectively, and $d$ is the dimension of the key vectors. This formulation necessitates the recomputation of attention weights at each layer, a computationally intensive task, particularly when the model is deployed in inference mode. To mitigate this, the concept of a KV-cache is employed, reducing the need to recompute $K$ and $V$ matrices for previously encountered tokens.

While prior methodologies have focused on sharing KV caches at different levels to minimize memory overhead, they predominantly operate under the assumption that attention weights differ significantly across layers, thereby necessitating individual computations to capture diverse contextual dependencies effectively. This assumption prompts a critical inquiry: Are the attention weights indeed markedly different across layers, or is this variation minimal enough to allow for a unified approach across multiple layers?

To explore this, we conducted an empirical analysis on the distribution of attention weights across different layers of the model. Based on the Llama2-7B-chat model, we processed the Massive Multitask Language Understanding (MMLU) dataset \cite{hendrycks2020measuring} to extract the attention matrices, $\text{softmax}(\frac{QK^T}{\sqrt{d}})$, for each layer. Given the variability in sequence lengths, we standardized these matrices to a uniform size by applying zero-padding to align them to a consistent shape of $\text{maxlen} \times \text{maxlen}$.

Our analysis employed the cosine similarity metric to compare the attention matrices of all layers, revealing a notable high degree of similarity across most of layers, particularly from indices 3 to 30. Contrastingly, the initial layers (0 and 1) and the final output layer (31) exhibited substantially lower similarity scores to middle layers. This observation is intuitive as the early layers are closer to the input token embeddings, requiring frequent adjustments to their attention distribution to accurately abstract semantic meanings from diverse inputs. Similarly, the final layer's unique role in predicting the next token justifies its distinct attention pattern.

Inspired by these findings, we hypothesize that the high similarity in attention weights across the majority of layers could allow for a shared representation of these weights, thus eliminating the need for separate softmax computations in each layer and reducing the key cache size. Such a strategy could not only streamline the inference process but also enhance computational efficiency significantly.

Based on the observed uniformity in attention weights, we propose a novel algorithm as shown in Algorithm \ref{alg:shared_attention}, \textit{Shared Attention}, which utilizes a single shared attention matrix across multiple layers. This approach fundamentally redefines the operational paradigm by maintaining a consistent attention mechanism across various contextual layers, thereby reducing redundancy and enhancing inference speed.

\begin{algorithm}[tb]
\caption{Shared Attention Algorithm}
\label{alg:shared_attention}
\textbf{Input}: Set of layers $L$, input tokens $X$\\
\textbf{Parameters}: Attention span $S$ (e.g., layers 23 to 30)\\
\textbf{Output}: Updated attention weights across specified layers
\begin{algorithmic}[1] 
\STATE Initialize attention weights $A \gets \emptyset$
\FOR{each layer $l \in S$}
    \IF{first layer in $S$}
        \STATE Compute initial attention weights $A_l \gets \text{softmax}(\frac{Q_l K_l^T}{\sqrt{d_k}})$
        \STATE Set $A \gets A_l$
    \ELSE
        \STATE Share attention weights $A_l \gets A$
    \ENDIF
    \STATE Apply shared attention to compute outputs $O_l \gets A_l \cdot V_l$
\ENDFOR
\STATE Adjust subsequent layers' inputs using outputs from $S$
\STATE Continue processing remaining layers with standard attention
\STATE \textbf{return} Final output after processing all layers
\end{algorithmic}
\end{algorithm}

\begin{figure*}[ht]
    \centering
    \includegraphics[width=\textwidth]{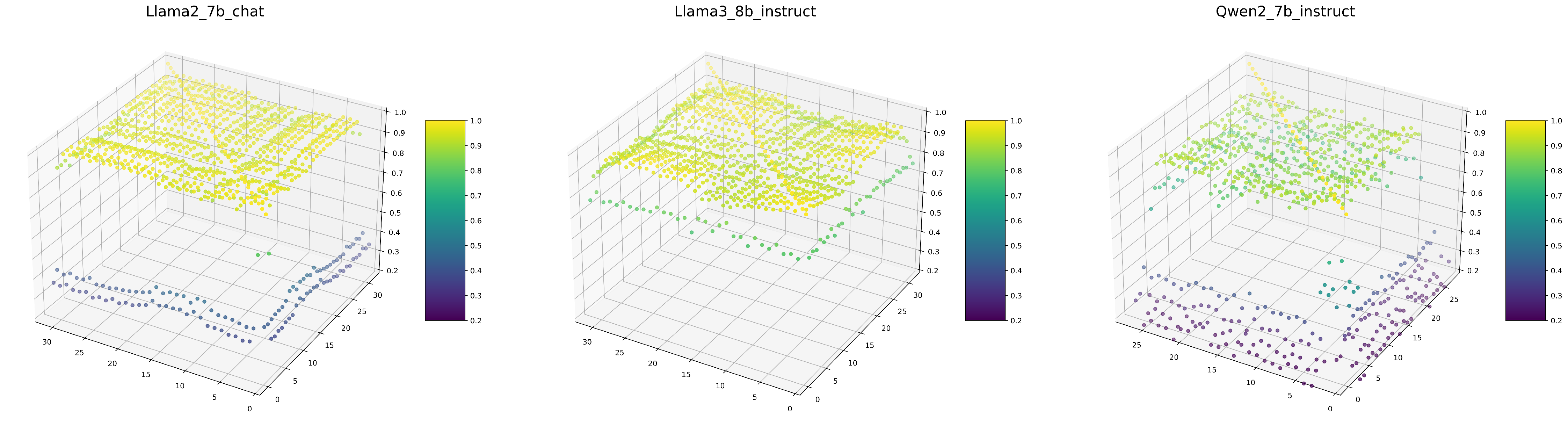}
    \caption{Layer-wise similarity of attention weights across various LLMs. The x-axis and y-axis represent the layer indices, while the z-axis depicts the cosine similarity values. The distinct similarity patterns are indicative of the specific functional roles each group of layers plays within the overall architecture.}
    \label{fig:attention_weights}
\end{figure*}

\subsection{Comparison with Existing Approaches}
The original self-attention mechanism in Transformers, characterized by the Multi-Head Attention (MHA) model, necessitates caching the keys ($K$) and values ($V$) in each head and layer to accelerate inference \cite{vaswani2017attention}. This requirement has historically imposed a significant memory overhead, prompting a series of innovations aimed at reducing this burden.

Among these, Multi-Query Attention (MQA) and its more generalized counterpart, Grouped-Query Attention (GQA), consolidate the KV cache by allowing multiple query heads within the same layer to share a singular set of K and V matrices. This approach effectively reduces the number of unique key and value pairs that must be stored and retrieved during the computation process. Subsequently, Cross-Layer Attention (CLA) extends this concept by facilitating the sharing of K and V matrices across different layers, thereby offering further reductions in the memory footprint required for KV storage.

Our method, however, introduces a fundamentally different paradigm in addressing the challenges of self-attention. While previous methods have focused on reducing the redundancy in storing K and V matrices, our approach centers on the optimization of the computation of attention weights themselves. In standard practice, the cached keys ($K$) are primarily utilized to compute attention weights in conjunction with the queries ($Q$). Instead of indirectly facilitating this interaction through shared KV matrices, our method proposes the direct sharing of the resultant attention weights—specifically, the softmax-normalized scores.

This not only diminishes the memory requirements by obviating the need to store separate sets of keys for each layer but also significantly reduces the computational complexity. By sharing the pre-computed softmax results across layers, our approach circumvents the repeated calculation of softmax, which is often one of the most computationally intensive operations in the attention mechanism. This efficiency gain is reflected in a substantial reduction in the number of floating-point operations (FLOPs) required during model inference, enhancing both the speed and scalability of Transformer deployments.

Unlike traditional methods that optimize memory use by sharing physical keys and values across layers or heads, our Shared Attention model innovates on the computational process itself, exploiting the consistent patterns in attention weights to streamline operations across multiple layers of the Transformer architecture.

\section{Isotropic Attention Distribution}
In an extensive analysis of layer-specific attention weights across a spectrum of LLMs, we explored the attention dynamics within models such as Llama2-7B-chat, Llama3-8B-instruct, Llama3-70B-instruct, Baichuan2-7B-chat, Qwen2-7B-instruct, and Qwen2-72B-instruct \cite{touvron2023llama, yang2023baichuan, bai2023qwen}. These models were evaluated using the MMLU.

Our investigations reveal a self-organization pattern in the attention weights across these diverse models. As depicted in Figure \ref{fig:attention_weights}, there exists a consistent global similarity pattern in the layers' attention weights across all tested models. This pattern suggests an inherent structural characteristic in the way LLMs process information, which can be broadly segmented into four distinct groups:

\begin{itemize}
    \item \textbf{Group 1:} Comprising the initial layers (indices 0 and 1), this group is situated closest to the input tokens and primarily focuses on abstracting token-level semantic information. These layers exhibit data-dependent attention patterns that are crucial for the initial semantic processing of the inputs.
    
    \item \textbf{Group 2:} This group includes layers immediately following the first group and extends up to layer index 5. Layers in this segment demonstrate high internal similarity in attention weights but are markedly different from those in other groups. These layers likely serve as transitional zones where intermediate semantic features are refined.
    
    \item \textbf{Group 3:} Encompassing layers post-Group 2 and extending to the penultimate layer, this is the largest group both in terms of the number of layers and their role within the architecture. The layers within this group display a high degree of similarity, suggesting an isotropy in the attention mechanism where the refined features are consistently utilized to inform the model's deeper contextual understanding.
    
    \item \textbf{Group 4:} The final group, consisting solely of the output layer, distinctively processes the aggregated contextual information to generate outputs. This layer's attention weights diverge from those observed in other layers, underscoring its specialized role in the final decision-making process.
\end{itemize}

The distinct attention weight patterns identified across these groups reinforce the concept of functional specialization within LLMs. This segmentation not only highlights the diverse roles of different layers in processing inputs but also supports the potential for optimizing computational strategies, such as our proposed Shared Attention method, by manipulating these inherent patterns to reduce computational redundancy.

\begin{figure*}[ht]
    \centering
    \includegraphics[width=\textwidth]{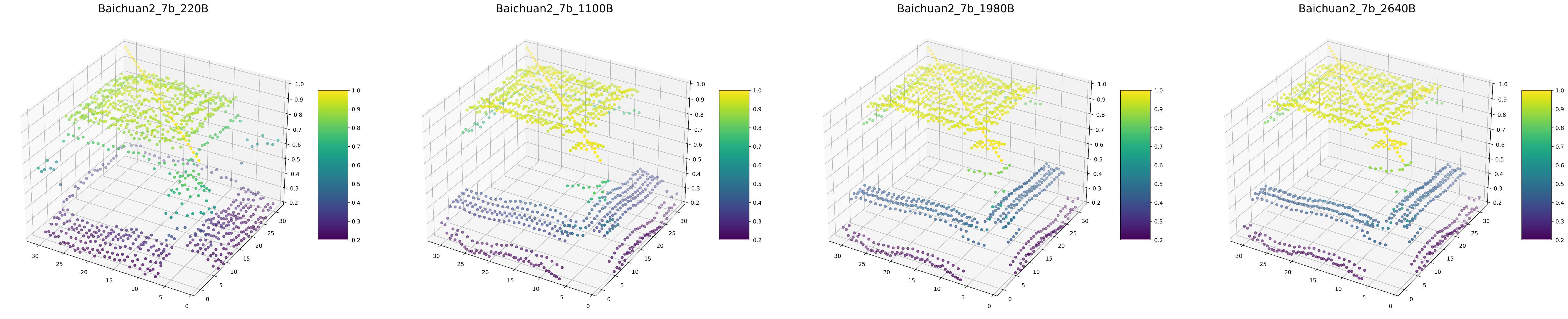}
    \caption{Evolution of layer attention weights similarity throughout the pretraining phase of the Baichuan2 7B model, as it processes trained tokens from 220 billion to 2.6 trillion. The color gradient in the visualization represents cosine similarity, effectively illustrating the transition in attention patterns from the initial to the advanced stages of pretraining.}
    \label{fig:pretraining_dynamics}
\end{figure*}

\subsection{Dynamics During Pretraining}

To elucidate the formation and evolution of attention weight patterns during the pretraining phase of LLMs, we utilized intermediate checkpoints of the Baichuan 7B model, provided by the model developers. These checkpoints, spanning from 0.2T to 2.6T tokens processed, offer a unique point of view to observe the dynamic shifts in attention mechanisms as the model gains exposure to an increasing volume of data.

We applied a consistent metric for measuring the similarity of attention weights across layers at each pretraining checkpoint. Additionally, the final chat model, fine-tuned to align with human reference responses, was included to benchmark the evolution against practical application outcomes. The dynamics of these attention weights are visualized in Figure \ref{fig:pretraining_dynamics}, which illustrates the progressive differentiation and stabilization of attention patterns across the model's layers.

As observed in the early pretraining stage at 0.2T tokens, Groups 1 and 2 appear merged, indicating a less differentiated processing strategy across these initial layers. This combination suggests that early in training, the model does not distinctly separate token-level semantic processing from intermediate semantic refinement. However, as the model progresses to 1.0T tokens, a clear division emerges between Groups 1 and 2. This separation aligns with the model beginning to form more specialized and efficient strategies for handling different types of information across its architecture.

The similarity within Group 3, which encompasses the bulk of the model's layers, shows a marked improvement from a similarity score of 0.8 to 0.9. This increase is indicative of the model's attention mechanism stabilizing and becoming more consistent in its approach to processing the bulk of contextual information.

The training advancements observed across the pretraining checkpoints not only demonstrate significant shifts in the internal structure of the model's attention mechanisms but also correlate positively with performance improvements on multiple benchmarks. This includes results on the MMLU, CMMLU \cite{li2023cmmlu}, and C-Eval \cite{huang2024c} 5-shot accuracy tests, which have reportedly improved from a baseline accuracy of 0.25 to 0.50 \cite{yang2023baichuan}. This notable enhancement underscores the intrinsic link between the refinement of attention mechanisms within LLMs and their enhanced capabilities in natural language understanding tasks.

Moreover, further examination of the model’s development, as observed in supplementary material, reveals that the similarity within Group 3—comprising the core contextual processing layers of the model—continues to enhance after the alignment stage. This observation suggests that the alignment process, typically aimed at fine-tuning the model to more closely mirror human-like understanding and response generation, also contributes to the stabilization of the model's attention mechanisms.

\section{Experiments and Discussion}
To validate the efficacy of our proposed Shared Attention (SA) method, we conducted series of experiments. These experiments were designed to test the robustness of SA under various configurations and to evaluate its performance on widely recognized benchmarks.

Initially, we applied the SA mechanism directly to advanced LLMs without any prior training to assess its impact on pre-trained models. This experiment aimed to understand the immediate effects of SA when integrated into existing model architectures. We evaluated the performance of these models on standard LLM benchmarks, including GLUE (General), GSM8k (Arithmetic), HellaSwag (Reasoning), and MMLU (Knowledge) \cite{wang2018glue, cobbe2021training, zellers2019hellaswag}. As anticipated, the direct application of SA resulted in a loss of accuracy on some benchmarks. This outcome is consistent with our expectations given the lack of retraining to adapt the models fully to the nuances of the Shared Attention mechanism. Due to computational constraints, it was impractical for our team to pretrain an LLM from scratch incorporating SA.

To further probe the capabilities of SA under a training regimen, we fine-tuned base LLMs equipped with Shared Attention on the publicly available Instruct dataset \cite{taori2023alpaca}. Post fine-tuning, these models were tested against the same benchmarks to find out any performance changes. This approach allowed us to measure the adaptability of SA when models are trained to accommodate its dynamics.

These experiments collectively demonstrate the potential of Shared Attention to modify the traditional attention mechanism in LLMs, showing a promising avenue for reducing computational demands while maintaining, and in some cases enhancing, model performance. The detailed results and further discussion on each benchmark and dataset are provided in the subsequent sections.

\subsection{Experimental Setup}
For the fine-tuning experiments, we utilized the Llama2-7B and Llama3-8B base models. These experiments were conducted on a robust hardware configuration consisting of two NVIDIA A100 80GB GPUs. Optimization of the models was carried out using the AdamW optimizer, with an initial learning rate set at $2 \times 10^{-5}$. We employed the bf16 datatype for model parameters, which enhances the numeric range and stability during backpropagation, crucial for maintaining precision in large model training.

Each GPU handled a micro-batch size of 16, leveraging gradient accumulation techniques to effectively manage the computational load. Additionally, we utilized DeepSpeed Zero Stage 3 to optimize the distribution of model and optimizer parameters and enhance memory management across the GPUs, ensuring efficient use of available resources. The fine-tuning process spanned two epochs and employed the standard Alpaca instruction format, which is designed to improve the responsiveness and accuracy of the models in handling instruction-based tasks.

\subsection{Direct Application of Shared Attention}
\begin{figure}[ht]
    \centering
    \includegraphics[width=0.4\textwidth]{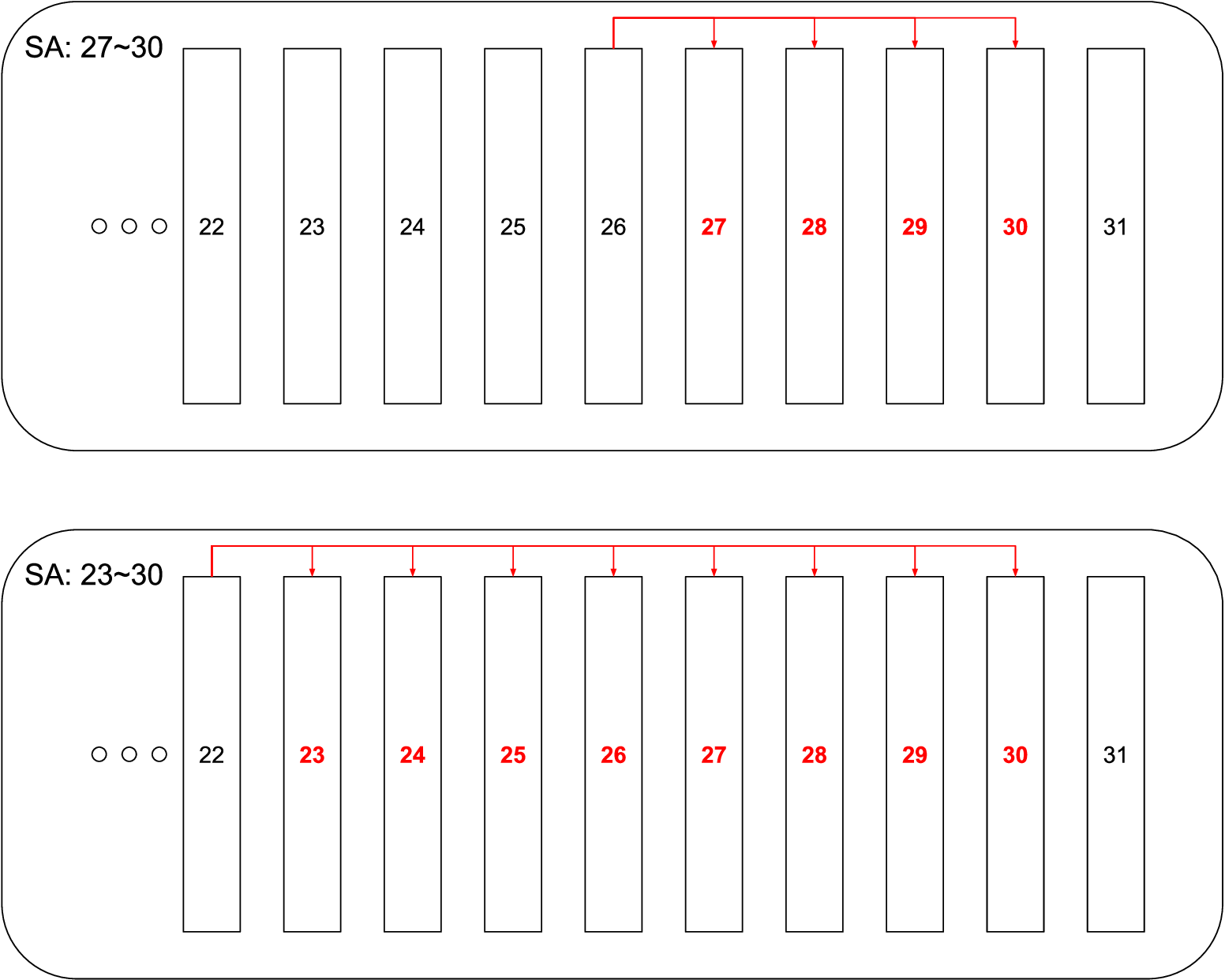}
    \caption{The figure illustrates the implementation of Shared Attention within specific layer segments of the model. Shared Attention spans from layer 27 to 30 for a four-layer segment and from layer 23 to 30 for an eight-layer segment.}
    \label{fig:sa_range}
\end{figure}

The application of SA was tested across discrete segments of layers within the Llama2-7B and Llama3-8B models, each comprising 32 layers in total. To evaluate the robustness and adaptability of SA as shown in Figure \ref{fig:sa_range}, it was implemented in varying layer segments, ranging from narrower spans such as four layers (e.g., SA:15$\sim$18) to broader spans such as eight layers (e.g., SA:23$\sim$30).

\begin{figure}[ht]
    \centering
    \includegraphics[width=0.4\textwidth]{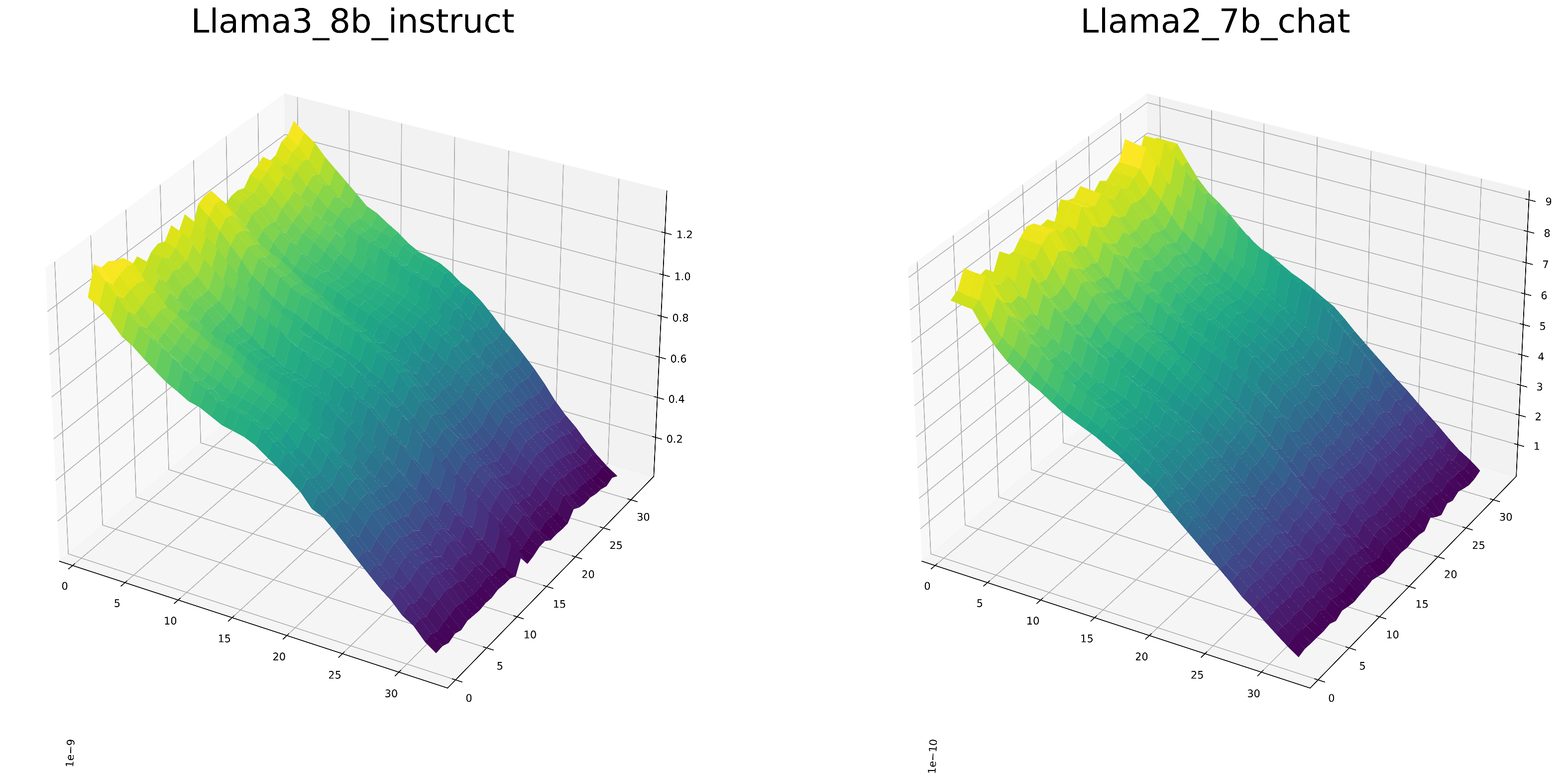}
    \caption{The figure displays the weighted cumulative variance for the Llama2-7B-chat and Llama3-8B-instruct models. The two lower axes represent the model's structure: the left axis details the 32 layers, and the right axis shows the 32 heads within each layer. The z-axis represents the variance values.}
    \label{fig:variance_layer_attention}
\end{figure}

Preliminary assessments of SA in the earlier layers of Llama2-7B (e.g., layers 3 to 6) resulted in an explosion of perplexity, indicating significant disruptions in the model's ability to predict subsequent tokens accurately. This phenomenon underscores the crucial role that attention score variances play in the model's early stages of processing, which are essential for initial context setting and feature extraction. To quantitatively assess the impact of attention variance throughout the model, we conducted a detailed variance analysis. We applied the same computational method used to obtain attention mean scores to calculate the variance of attention weights in Llama2-7B and Llama3-8B while processing the MMLU dataset. We further explored the potential influence of attention variance in downstream layers by computing a weighted cumulative variance. This metric aggregates the variances of all downstream layers starting from each specific layer, weighted by the average of these summed variances. As illustrated in Figure \ref{fig:variance_layer_attention}, the analysis revealed that early layers exhibited significantly higher weighted variances compared to latter layers. This variance tends to decrease as one progresses through the model's architecture, suggesting a stabilization of attention mechanisms in the latter layers. Given these results, our experiments predominantly focused on the application of SA in the latter layers, where such variances appear to stabilize.

\begin{table*}[htb]
\centering
\begin{tabularx}{\textwidth}{@{}cccccc@{}}
\toprule
Model & GLUE & \begin{tabular}[c]{@{}c@{}}GSM8K \\ 5-shot\end{tabular} & HellaSwag & MMLU \\
\midrule
Llama2-7B & 0.4050 $\pm$ 0.0019 & 0.1395 $\pm$ 0.0095 & 0.5713 $\pm$ 0.0049 & 0.4119 $\pm$ 0.0041 \\
Llama2-7B$_{\text{SA}:23\sim30}$ & 0.3819 $\pm$ 0.0019 & 0.0728 $\pm$ 0.0072 & 0.5575 $\pm$ 0.0050 & 0.3794 $\pm$ 0.0040 \\
Llama2-7B$_{\text{SA}:27\sim30}$ & 0.3882 $\pm$ 0.0019 & 0.1243 $\pm$ 0.0091 & 0.5616 $\pm$ 0.0050 & 0.4056 $\pm$ 0.0041 \\
Llama2-7B$_{\text{SA}:23\sim26}$ & 0.4351 $\pm$ 0.0019 & 0.1122 $\pm$ 0.0087 & 0.5681 $\pm$ 0.0049 & 0.3994 $\pm$ 0.0040 \\
Llama2-7B$_{\text{SA}:19\sim22}$ & 0.3996 $\pm$ 0.0019 & 0.0834 $\pm$ 0.0076 & 0.5553 $\pm$ 0.0050 & 0.3926 $\pm$ 0.0040 \\
Llama2-7B$_{\text{SA}:15\sim18}$ & 0.3731 $\pm$ 0.0019 & 0.0220 $\pm$ 0.0040 & 0.4790 $\pm$ 0.0050 & 0.3378 $\pm$ 0.0047 \\
\midrule
Llama2-7B-Instruct-SFT & 0.5372 $\pm$ 0.0019 & 0.1440 $\pm$ 0.0097 & 0.5772 $\pm$ 0.0049 & 0.3722 $\pm$ 0.0040 \\
Llama2-7B-Instruct-SFT$_{\text{SA}:23\sim30}$ & 0.5401 $\pm$ 0.0019 & 0.0758 $\pm$ 0.0073 & 0.5671 $\pm$ 0.0049 & 0.3717 $\pm$ 0.0040 \\
\midrule
Llama3-8B & 0.4804 $\pm$ 0.0019 & \bf{0.5155} $\pm$ 0.0138 & 0.6009 $\pm$ 0.0049 & \bf{0.6198} $\pm$ 0.0038 \\
Llama3-8B$_{\text{SA}:23\sim30}$ & \bf{0.5595} $\pm$ 0.0019 & 0.3275 $\pm$ 0.0129 & 0.6011 $\pm$ 0.0049 & 0.6122 $\pm$ 0.0038 \\
Llama3-8B$_{\text{SA}:27\sim30}$ & 0.5532 $\pm$ 0.0019 & 0.4526 $\pm$ 0.0137 & \bf{0.6060} $\pm$ 0.0049 & 0.6163 $\pm$ 0.0038 \\
Llama3-8B$_{\text{SA}:23\sim26}$ & 0.5024 $\pm$ 0.0019 & 0.4556 $\pm$ 0.0137 & 0.5993 $\pm$ 0.0049 & 0.6189 $\pm$ 0.0038 \\
Llama3-8B$_{\text{SA}:19\sim22}$ & 0.5115 $\pm$ 0.0019 & 0.3745 $\pm$ 0.0133 & 0.5829 $\pm$ 0.0049 & 0.6181 $\pm$ 0.0038 \\
Llama3-8B$_{\text{SA}:15\sim18}$ & 0.4685 $\pm$ 0.0019 & 0.0136 $\pm$ 0.0032 & 0.5307 $\pm$ 0.0050 & 0.3019 $\pm$ 0.0038 \\
\bottomrule
\end{tabularx}
\caption{Performance metrics for different models across tasks}
\label{tab:performance_metrics}
\end{table*}

The outcomes of these experiments, as summarized in Table \ref{tab:performance_metrics}, reveal interesting patterns. For the Llama2-7B model, implementing SA in the latter layers (e.g., SA:23$\sim$26 and SA:27$\sim$30) maintained relatively stable performance across a variety of benchmarks, including GLUE and MMLU. Conversely, extending the scope of SA to encompass more layers, particularly mid-level layers such as SA:15$\sim$18, led to a noticeable degradation in tasks requiring mathematical reasoning (GSM8K).

In comparison, the Llama3-8B model, which inherently showed higher layer-wise attention similarity as discussed in the previous sections, exhibited less performance deterioration when SA was applied. After implementing SA in the layers closer to the model's output (e.g., SA:27$\sim$30), the Llama3-8B even outperformed its original configuration on the GLUE benchmark, suggesting that strategic placement of SA can potentially enhance the model's performance in complex natural language understanding tasks.

\subsection{Fine-Tuning on Instruct Dataset}
Given the computational constraints that preclude the pretraining of LLMs with SA from scratch, we adopted to fine-tune existing LLMs to evaluate whether fine-tuning could ameliorate the performance deficits observed with the direct application of SA. This approach was particularly aimed at understanding the adaptability of SA under a more controlled learning regimen.

Fine-tuning was conducted on the publicly available Instruct dataset, which is designed to evaluate models on tasks that require following complex instructions. This dataset was chosen because it challenges the models to utilize their learned representations effectively, making it an ideal benchmark for testing the efficacy of modifications like SA.

The results, as summarized in Table \ref{tab:performance_metrics}, demonstrate a narrowed performance gap between the original models and those modified with SA. For instance, while the original Llama2-7B model outperformed the SA version in direct application tests, the fine-tuned Llama2-7B$_{\text{SA}:23\sim30}$ showed significant improvements across multiple metrics. This suggests that fine-tuning enables the model to better integrate and leverage the Shared Attention mechanism, effectively regaining some of the lost performance noted in the initial application of SA.

These findings indicate the potential of fine-tuning as a viable method for integrating new architectural changes like SA into existing models. The recovery in performance indicates that with adequate training, the initial disadvantages of directly applying SA can be mitigated, leading to enhanced model capabilities that more closely align with or even exceed their original configurations.

\subsection{Future Directions}
Our experimental investigations have demonstrated that implementing Shared Attention (SA) across multiple latter layers in LLMs arouses minimal accuracy loss, making it a promising approach for enhancing model efficiency. Furthermore, our analysis reveals a trend towards isotropic attention patterns during the pretraining process, indicating that the models' attention mechanisms tend to stabilize as they process more data.

Given these insights, integrating SA from the pretraining appears to be a particularly beneficial strategy. This early integration could allow models to better adapt to the streamlined attention mechanism, potentially improving performance and efficiency across various tasks. The foundational embedding of SA might simplify later adaptations and inherently supports efficient attention dynamics.

Another promising research direction involves exploring combinations between SA and other attention-sharing strategies like Cross-Layer Attention (CLA). Combining SA with methods such as CLA could exploit the strengths of both approaches, leading to a more robust and flexible attention mechanism. This holistic approach to attention management could provide a comprehensive solution that maximizes both computational efficiency and model scalability.

By pursuing these avenues, future research can not only refine the application of Shared Attention within LLMs but also explore its full potential in enhancing the architectural and operational efficiency of next-generation language models. These efforts could lead to models that are better equipped to handle the increasing complexity and diversity of tasks in natural language processing.

\section{Related Work}
Efficient memory management in transformers is a critical area of research with diverse objectives ranging from reducing memory bandwidth and storage requirements to optimizing computational costs during both training and inference phases. Notably, our work focuses on minimizing the size of the inference Key-Value (KV) cache that persists between model passes, thereby enhancing model efficiency without a significant compromise in performance.

\subsection{Memory Efficiency in Attention Mechanisms}
Significant efforts have been made to address the efficiency of the KV cache post-training. Techniques such as KV cache compression have been explored extensively. For instance, methods like KVQuant \cite{hooper2024kvquant} and KIVI \cite{liu2024tuning} employ quantization strategies to reduce the memory footprint of KV pairs to just a few bits. Moreover, works such as AttentionSink \cite{xiao2023efficient} and Scissorhands \cite{liu2024scissorhands} introduce sparsity into the KV cache by selectively storing elements based on their proximity or importance to the generation token, thus reducing the overall storage requirements.

\subsection{Architectural Innovations for Reducing KV Cache}
Architectural modifications aimed at reducing the KV cache size are pivotal in enhancing the efficiency of large language models. Such strategies include limiting the effective sequence length, as seen in Sparse Attention \cite{child2019generating}, which constrain attention to local windows to reduce both computational load and memory overhead. Another approach involves replacing traditional softmax attention with scalable alternatives like linear attention \cite{katharopoulos2020transformers}, which maintains constant space complexity and offers more graceful scaling with respect to the token count. Additionally, methods such as Grouped-Query Attention (GQA) \cite{ainslie2023gqa} and Multi-Query Attention (MQA) \cite{shazeer2019fast} aggregate attention across multiple queries, significantly decreasing the memory footprint by sharing KV pairs across attention heads. These innovations collectively contribute to reducing the redundancy in attention calculations and are directly relevant to our work, informing our development of the Shared Attention mechanism that further optimizes memory usage by sharing attention weights across layers.

\section{Conclusion}
In this paper, we explored the attention dynamics within advanced LLMs and observed that the attention distribution across layers tends to isotropize following extensive pretraining. This isotropic pattern of attention, where layers exhibit similar attention mechanisms, inspired a novel approach to attention sharing that departs from conventional methods.

Traditionally, methods like MQA and CLA have focused on sharing KV caches to reduce memory overheads but still required the computation of attention weights independently across each layer. Our proposed Shared Attention (SA) method bypasses this redundancy by directly sharing the computed attention weights across multiple layers. This approach not only significantly reduces the size of the KV cache but also decreases the computational FLOPs required during model inference.

The introduction of Shared Attention represents a paradigm shift in the design of attention mechanisms in neural networks, emphasizing efficiency without compromising the model's performance. By reducing both the computational burden and memory requirements, SA enables more scalable and efficient deployment of LLMs, particularly in environments where resources are constrained.

This research paves the way for further explorations into efficient model architectures and opens up new possibilities for the application of LLMs across a broader spectrum of tasks and datasets. Future work will focus on expanding the applicability of Shared Attention, exploring its integration during the initial phases of model training, and combining it with other optimization techniques to maximize the operational efficiency of LLMs.

\bibliography{aaai25}

\end{document}